# Multilingual Controlled Generation And Gold-Standard-Agnostic Evaluation of Code-Mixed Sentences

**Ayushman Gupta**[*1]    **Akhil Bhogal**[*2]    **Kripabandhu Ghosh**[2]

[1]South City International School, India   [2]IISER Kolkata, India

ayushman2346@gmail.com  akhilbhogal25@gmail.com  kripaghosh@iiserkol.ac.in

## Abstract

Code-mixing, the practice of alternating between two or more languages in an utterance, is a common phenomenon in multilingual communities. Due to the colloquial nature of code-mixing, there is no singular correct way to translate an English sentence into a code-mixed sentence. For this reason, standard n-gram-based MT evaluation metrics such as the BLEU score are not appropriate for code-mixed evaluation. To demonstrate this, we propose a novel method for code-mixed text generation: Controlled Generation, which parameterizes the code-mixing degree (CMD) and enables the generation of multiple semantically equivalent code-mixed sentences from a given English sentence. We introduce a robust new evaluation metric: GAME: *A Gold-Standard Agnostic Measure for Evaluation of Code-Mixed Sentences*. GAME is both language-agnostic and gold-standard-agnostic, i.e. unlike other metrics, GAME does not require gold-standard code-mixed sentences for evaluation, thus eliminating the need for human annotators in the code-mixed evaluation process. When used to evaluate semantically equivalent code-mixed sentences, we find that GAME scores have a lower standard deviation than BLEU scores. Further, we create and release a dataset containing gold-standard code-mixed sentences across 4 language pairs: English-{Hindi, Bengali, French, Spanish} to encourage more computational research on code-mixing.

## 1 Introduction

Code-mixing, or code-switching, refers to the practice of alternating between two or more languages in a single utterance (Poplack, 2001). This is commonly observed in bilingual and multilingual communities where speakers are fluent in two or more languages. Code-mixing involves a 'matrix' language, which influences the grammar of the sentence, and an 'embedded' language, from which words or phrases are inserted into a monolingual sentence to form a code-mixed sentence. Moreover, code-mixing is the language of social media (Chung et al., 2022). In multilingual countries such as India, code-mixing is commonplace: a significant number of Indians who are fluent in both English and Hindi tend to speak 'Hinglish' rather than English or Hindi, in informal settings. It is for this reason that the phenomenon of code-mixing has been of great interest in NLP.

The premise of our work (as depicted in Figure 1) is that an English sentence often has multiple *semantically equivalent*, equally valid code-mixed translations. Therefore, a robust code-mixed evaluation metric must assign equal and perfect scores to all these sentences. However, popular MT evaluation metrics such as the BLEU score (Papineni et al., 2002a) use n-grams to measure the similarity between a reference and a candidate sentence which renders them inappropriate for code-mixed evaluation (Srivastava and Singh, 2021).

To address the abovementioned gaps in literature, in this paper we propose a novel method to generate semantically equivalent code-mixed sentences from a given English sentence by parameterizing the Code-Mixing Degree (CMD). We call this method **Controlled Generation** (CG) (see Section 4). CG also attempts emulate real-world code-mixing by aligning its generations to code-mixing trends seen in social media data. Furthermore, as an alternative to BLEU, we introduce a novel, robust code-mixed evaluation metric, **GAME**: A **G**old-Standard **A**gnostic **M**easure for **E**valuation of Code-Mixed Sentences (see Section 5). *To the best of our knowledge, GAME marks the first attempt at creating a pipeline to automatically evaluate code-mixed generations.* Despite the ubiquity of code-mixing, gold-standard code-mixed data, especially for low-resource languages, is scarce. This paucity of data stems from the fact that the creation of gold-standard code-mixed data requires significant human annotation efforts, which can be expensive. GAME is gold-standard-agnostic, i.e. unlike other evaluation metrics, it does not require human-generated gold-standard code-mixed sentences for evaluation. Although we explore Controlled Generation and test GAME in four language pairs, i.e. English-{Hindi, Bengali, French, Spanish}, Controlled Generation can easily be extended to other language pairs, and GAME is language-agnostic.

---

[*]Equal contribution.

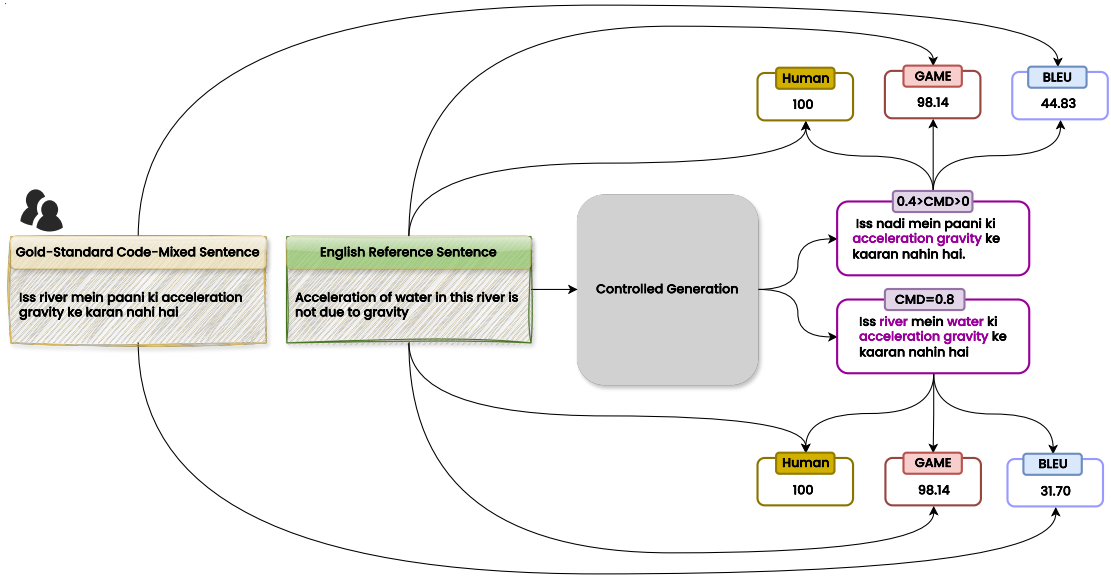

Figure 1: A depiction of our work: generation and evaluation of two semantically equivalent code-mixed English-Hindi sentences using Controlled Generation with two different degrees of code-mix, viz. CMD (see Section 4 for more details) and GAME (see Section 5 for more details) respectively

Interestingly, CG and GAME beautifully complement each other to introduce a novel evaluation paradigm. As shown in Figure 1, we use CG to generate two semantically equivalent code-mixed (English-Hindi) translations of the English sentence: *"Acceleration of water in this river is not due to gravity"*. We requested human experts to evaluate these translations, and then evaluate them using the BLEU score (Papineni et al., 2002a), and GAME. We find that both the code-mixed sentences generated using CG are high quality generations as they received a perfect human score of 100. However the BLEU scores for these sentences (44.83 and 31.70) are unequal and not aligned with the human score. On the other hand, the GAME scores for these sentences (98.14) are equal and more representative of the human score. Additionally, a gold-standard code-mixed sentence had to be created by human annotators to facilitate evaluation using BLEU, while GAME required only the English reference sentence (see Section 5.2 for more details).

In order to encourage more computational research on code-mixing and contribute towards remedying the scarcity of gold-standard code-mixed data, we release a dataset[1] containing **1506** gold-standard English-Hindi, English-Bengali, English-French, and English-Spanish code-mixed sentences (see Section 3).

To summarize, the contribution of the paper is three-fold. Firstly, we propose a novel scheme for automatically generating semantically equivalent code-mixed generations parameterizing the Code-Mixing Degree, viz. CG (see Section 4). Secondly, we propose a novel human-agnostic and language-agnostic measure, viz. GAME, for evaluation of code-mix generations (see Section 5). Finally, we create a code-mixed dataset containing 1.5k gold-standard code-mixed sentences across four language pairs: English-Hindi, English-Bengali, English-French, and English-Spanish (see Section 3) for evaluation of the proposed schemes.

## 2 Related Work

**Code-Mixed Text Generation** Gupta et al. 2020 present a semi-supervised approach to generate code-mixed text using a pre-trained encoder and transfer learning for diverse eight language pairs. Hsu et al. 2023 introduces GLOSS, a model that synthesizes code-switched text for language pairs not present in training data, leveraging a pre-trained multilingual machine translation model with an additional code-switching module. Gautam et al. 2021 fine-tunes mBART for Hinglish generation and also utilizes the pre-training of model in Devanagri script.

---
[1]The dataset can be found at https://rb.gy/i7u987

**Code-Mixed Text Generation Using LLMs** Zhang et al. 2023 analyze the code-switching abilities of multilingual LLMs across 4 tasks, including Machine Translation (MT). They perform 0-shot prompting on several LLMs to translate English sentences to Hinglish, and vice versa.

**Controlled Generation** Mondal et al. 2022 is the only other work to our knowledge that focuses on controlling the degree of code-mixing. They present CoCoa, an encoder-decoder translation model, for this purpose. While CoCoa tries to have control over the Code-Mixing Index (CMI) which relies on the knowledge of the language down to the individual tokens, our approach allows one to exercise control in choosing the matrix language and embedded language while systematically choosing the switch points, by prompting an LLM and using a Twitter code-mixed dataset to emulate real world code-mixing.

**Automatic evaluation** Garg et al. 2021 introduce MIPE: A Metric Independent Pipeline for Effective Code-Mixed NLG Evaluation. While MIPE measures the quality of a code-mixed sentence by adjusting for linguistic issues such as spelling variations, language switching, missing words, and scarcity of gold-standard code-mixed sentences, GAME is a gold-standard independent pipeline measures the quality of a code-mixed translation based on an English reference sentence.

## 3 Dataset Creation

There is a scarcity of publicly available *gold-standard* parallel code-mixed datasets. The availability of high-quality, noise-free parallel code-mixed datasets is essential for tasks such as training code-mixed evaluation metrics. Several English-Hindi (Dhar et al., 2018; Parekh et al., 2020; Gautam et al., 2021), English-Bengali (Patra et al., 2018), English-French (Carpuat, 2014), and English-Spanish (Solorio and Liu, 2008; Ahn et al., 2020) datasets have been released in previous work. However, these datasets are either not parallel, not publicly available, or are Twitter datasets containing substantial noise in the form of misspellings, typos, and textisms. Gupta et al. 2020 release code-mixed datasets spanning eight language pairs. However, this code-mixed data is synthetically-generated and contains substantial noise. To overcome the lack of gold-standard Hinglish data, Srivastava and Singh 2021 introduce HinGE: a parallel dataset which contains English and human-generated Hinglish sentences.

While the HinGE dataset significantly contributes to the amount of publicly available gold-standard parallel Hinglish data, high-quality parallel code-mixed datasets for other language pairs remain scarce. Therefore, in order to enable more computational research on code-mixing, we create four code-mixed datasets, i.e. for the English-{Hindi, Bengali, French, Spanish} language pairs. Each dataset contains English sentences and a gold-standard code-mixed sentence corresponding to each English sentence. Our English-Hindi, English-Bengali, English-French, and English-Spanish datasets contain 370, 541, 248, and 347 sentences respectively, i.e. 1506 code-mixed sentences in all. The English sentences were chosen randomly from the dataset provided in Gupta et al. 2020 for 4 language pairs, viz. English-Hindi, English-Bengali, English-French, and English-Spanish. Some of the gold-standard sentences were created by cleaning and correcting Twitter data (Dhar et al., 2018; Aguilar et al., 2020; Patra et al., 2018). For the English-Hindi and English-Bengali language pairs, we represent the Hindi and Bengali words, respectively, in the Roman script. Table 7 presents examples of English sentences alongside the corresponding gold-standard code-mixed sentence created for each language pair. The datasets and the annotation process have been discussed in more detail in Appendix (D.1). To our knowledge, ours is the first dataset containing parallel gold-standard code-mixed data for multiple language pairs.

## 4 Controlled Generation (CG)

As depicted in Figure 1, multiple semantically equivalent code-mixed sentences may be generated from a given English sentence, whose Code-Mixing Degrees may vary according to the author's preference. With this in mind, we propose Controlled Generation: a novel method to generate code-mixed sentences, which not only al-

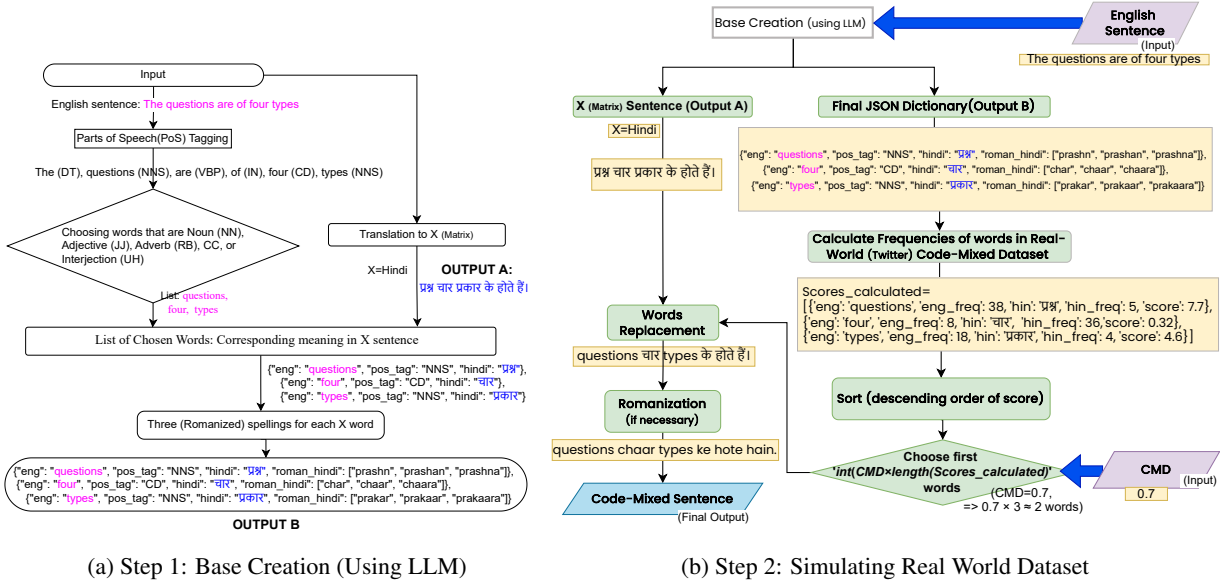

(a) Step 1: Base Creation (Using LLM)  (b) Step 2: Simulating Real World Dataset

Figure 2: Working of Controlled Generation. (X or Matrix Language is Hindi in the Example)

lows for control over the Code-Mixing Degree and the choice of the matrix language but also tries to emulate real-world code-mixing. A practical application of this algorithm may be found in a code-mixed chatbot, where controlling the Code-Mixing Degree would be crucial in tailoring the generations to the user's preferences.

Controlled Generation parameterizes the Code-Mixing Degree (CMD) of the generation, where $CMD \in [0, 1]$, thus allowing one to control the embedded-language contribution in the generation. A CMD value of 0 means that no code-mixing takes place, i.e. the generation is entirely in the matrix language. Controlled Generation allows one to choose the matrix and embedded languages. When Hindi is chosen as the matrix language, a value of 1 does not result in the generation of a monolingual English sentence. Conversely, if English is chosen as the matrix language (A.5), setting CMD to 0 will return in a monolingual English sentence.

### 4.1 Real-World Datasets

We try to simulate a real-life dataset in Controlled Generation. For this purpose, L3Cube-HingLID dataset[2], has been used for English-Hindi, Patra et al. 2018 for English-Bengali and LinCE (Aguilar et al., 2020) dataset for English-Spanish. Say, we have an English token "impossible" and its corresponding Hindi token "asambhav". Whether to use "impossible" or "asambhav" is determined by what is the trend or in other words what people like to use the most. This has been done by relative word frequency from the Real-World Datasets which will be explained.

CG requires a dataset but it is to be noted that this is not a limitation but a feature of CG. We can still generate sentences along with the aforementioned control without such a dataset.

### 4.2 CG with English as Embedded Language

Algorithm 1 (in Appendix E) explains Controlled Generation through pseudocode.

Figure 2 describes the working of Controlled generation accompanied by an English-Hindi generation example based on the input English sentence: *The questions are of four types*. The process is divided into two main steps:

#### 4.2.1 Step 1: Base Creation

This step is done by prompting with LLM. English sentence (Input) is first translated to the Matrix Language which in the example in the flowchart is Hindi ("प्रश्न चार प्रकार के होते हैं"). Through the same prompt, we obtain a list $W = \{w_1, w_2, \ldots, w_n\}$ of the replaceable words. These English words are chosen according to their PoS tags i.e. we choose words with specific tags as in flowchart only. This choice generates grammatically correct code-mixed sentences for all the four language pairs. We also obtain the corresponding 'switch points' where these words are to be replaced in the matrix language sentence by simply asking the LLM to find the corresponding meaning of these words in the translated sentence.Then, We ask the LLM to provide three Roman transliterations of each of these switch points or words. This is necessary for the next step because the Real-World dataset that we refer to for trend may contain spelling variations.

---
[2]https://github.com/l3cube-pune/code-mixed-nlp

Refer A.1 for the details of prompts and models used for this purpose.

### 4.2.2 Step 2: Simulating Real-World Dataset

To decide which words are to be replaced with their English translations based on the value of CMD, we refer to the trend of occurrence of words in real-world code-mixing. We construct a vocabulary of unique words from the dataset. Take the case of English-Hindi. Table 1 displays an excerpt from vocabulary created using English-Hindi dataset where $f(en)$ and $f(hi)$ are frequencies (counts) of the English and Hindi terms, respectively. The idea is that if the Hindi term for a word occurs less frequently in the dataset, replacement with the English term for that word should be preferred. As such, we calculate a 'score' $s$ for each replaceable word:

$$s = \frac{f(en)}{f(hi)}, \qquad f(hi) = \sum_{i=1}^{3} f(var_i)$$

where $f(var)$ is a variation of the Hindi term's Roman transliteration. Three variations are considered to account for differences in spelling. If $f(hi) = 0$, we assign a value of $inf$ to $s$.

| English Term | f(en) | Hindi Term | f(hi) |
|---|---|---|---|
| impossible | 15 | asambhav | 1 |
| water | 28 | pani, paani | 89, 101 |

Table 1: Words and their calculated frequencies

According to the value of the CMD parameter, the fraction of words to be replaced is chosen from the list $W$, with higher scores $s$ prioritized. These are replaced in the Hindi (Matrix Language) sentence to get a code-mixed sentence. In the flowchart, we have $CMD = 0.7$. Since there are three replaceable words, we replace $3 \times 0.7 \approx 2$ words. $f(en) > f(hi)$ for the words 'questions' and 'types', i.e., the value of $s$ is highest for these words, and they replace their Hindi terms 'प्रश्न' and 'प्रकार' in the Hindi sentence to give the code-mixed sentence: "questions चार types के होते हैं।". The output is the Romanization of this sentence: "questions chaar types ke hote hain".

**Replacement of all 'inf' score words: a subjective choice** In Controlled Generation, we encounter some matrix language words that are so rare that they do not appear in the dataset at all, resulting in $s = inf$. In such cases, if CMD is non-zero, all such words are replaced (or switched) first, even if their quantity exceeds the number of replacements allowed by input CMD. Then, if the CMD allows for further replacements, words with higher finite scores $s$ are considered next. This choice though, being subjective is grounded in the observation that individuals who do not use or know one such rare word are likely unfamiliar with other rare terms as well. This is especially true for English-Hindi and English-Bengali. Therefore, this choice is in favour of real-world code-mixing.

For sentences with **English as Matrix Language**, replace 'X' language words in English sentence with lower 's' prioritized instead. Refer A.5 for details

### 4.3 English-Hindi Specific CG

Hindi exhibit rich inflection, particularly in verbs. The verbs in Hindi convey information about gender along with other aspects and thus, simply replacing them with their English counterpart can result in low-quality (grammatically incorrect) code-mixed sentences. For instance, consider the English sentence 'He plays'. Its Hindi translation is *'Vaha khelta hai'*. Here, 'khelta' is the verb and thus, if language-agnostic CG is used, we get *'Vaha plays hai'*, which is a grammatically incorrect code-mixed sentence. The correct code-mixed sentence is *'Vaha play karta hai'*. In this sentence, the word *'karta'* is added after the root of the English verb 'play' to form a conjunct verb, i.e. 'play karta', based on the suffix *'ta'* in the word *'khelta'*. Thus, we introduce a modified prompt and verb-specific rules which allow us to generate grammatically correct code-mixed sentences. The detailed algorithm has been discussed in the Appendix A.2. In the following example, करने and करें are the added words:

English: *Select the line to invert.*
Hindi: रेखा को उलटने के लिए चुनें
Code-Mixed: *Line* को *invert* **करने** के लिए *select* **करें**

### 4.4 Critical Analysis of CG

We have evaluated CG using the BLEU score in section A.3 for the sake of some comparison with other works. Although we have computed BLEU

| English Sentence | Generations | GAME |
|---|---|---|
| The Congress should remove such differences, not create them. | Congress ko such antar mitaane chahiye, nahin banaane chahiye. | 96.23 |
| | Congress ko such differences mitaane chahiye, nahin banaane chahiye. | 96.23 |
| | Congress ko such differences mitaane chahiye, not banaane chahiye. | 96.23 |
| You have priceless batteries and an atomic bomb in your bag. | Aponar byage priceless batteries and ekta paromanobik boma royeche. | 88.37 |
| | Aponar byage priceless batteries and ekta atomic boma royeche. | 92.93 |
| | Aponar byage priceless batteries and ekta atomic bomb royeche. | 88.37 |
| This report is intended to facilitate this. | Este informe está destinado a facilitar esto. | 100.0 |
| | Este report está destinado a facilitar esto. | 100.0 |
| | Este report está destinado a facilitate esto. | 100.0 |

Table 2: English-X sentences generated through Controlled Generation and evaluated using GAME (5)

score, we note that according to previous work (Srivastava and Singh 2021, Gautam et al. 2021) and as has been shown in this paper (5.2.1), BLEU is inappropriate and ineffective for code-mixed evaluation. Thus, we evaluate CG using GAME 5, an evaluation metric tailored to code-mixed evaluation. Table 2 shows some examples of this, while a detailed analysis is presented in Section 5.2.1.

Generally, the generated sentences are grammatically correct and adhere to the specified CMD. Since CG involves multiple steps and relies on an LLM, errors, while being rare and minor, are observed. For example, in the third sentence in Table 2, the word 'ne' should not be present to ensure full grammatical correctness. This is a minor error that arises in the translation step. We have tested English-Hindi specific CG (6) for 100 sentences and observed that it eliminated the previous issues. Thus, CG, while being a rule-based method, results in generation of correct and real-world like code-mixed sentences.

## 5 GAME

As discussed previously (Figure 1) in the Introduction, the evaluation of code-mixed translations remains a challenge because of two key reasons:

**1)** Gold-standard code-mixed data is scarce. This follows from the fact that the creation of a corpus of gold-standard code-mixed data, especially for low-resource languages, is a laborious task and demands significant human annotation efforts.

**2)** For code-mixed evaluation, commonly-used Machine Translation (MT) evaluation metrics such as the BLEU score (Papineni et al., 2002b) require gold-standard code-mixed data, which is scarce, and are ineffective (Srivastava and Singh, 2021). The reason for their inefficacy is that an English sentence often has many equally correct, semantically equivalent code-mixed translations. These sentences typically differ in their Code-Mixing Degrees, and their choice of matrix language and words. Tables 2 and 3 show examples of multiple semantically equivalent code-mixed sentences generated from the same English sentence.

Recognizing these challenges, as well as the need for an automatic and efficient way to evaluate code-mixed translations, we present **GAME**: *A Gold-standard Agnostic Measure for the Evaluation of Code-Mixed Sentences*. To the best of our knowledge, GAME marks the first attempt at cre-ating a pipeline to automatically evaluate code-mixed translations.

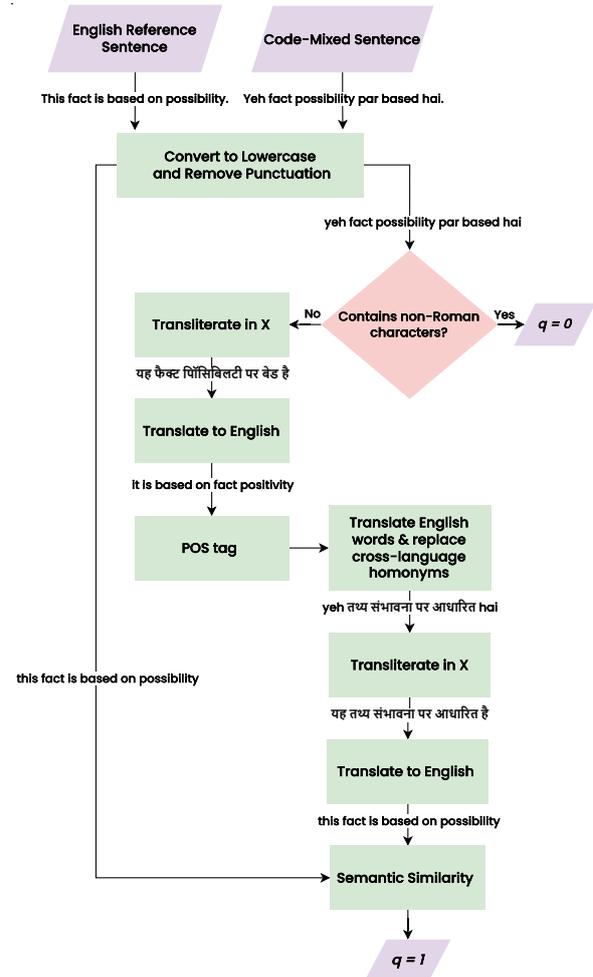

Figure 3: Flowchart for GAME

GAME, unlike CG (4) does not consider any trends for evaluation, and serves as an objective way to evaluate code-mixed sentences, i.e. high scores are assigned to grammatically correct code-mixed sentences that accurately convey the meaning of the English reference sentence.

Furthermore, unlike standard evaluation metrics such as BLEU, GAME does not require gold-standard code-mixed data for evaluation. Therefore, it eliminates the need for human annotation in order to create gold-standard code-mixed data for evaluation of code-mixed generations. It should be noted that although in this paper we implement GAME for 4 language pairs, it can also be used for any other English-X language pair.

Algorithm 2 (in Appendix F) explains GAME through pseudocode.

## 5.1 Process of Evaluation

At a high level, for a given English reference sentence $s_r$ and a code-mixed candidate sentence $s_{cm}$, GAME returns a score $q$, where $q \in [-1, 1]$. $q = 1$ implies that $s_{cm}$ is a perfect English-X translation of $s_r$. $q = 0$ when $s_{cm}$ contains Non-Roman characters.

Figure 3 shows stepwise with an English-Hindi example how a code-mixed sentence is evaluated using GAME. In this example, $s_r$ = "This fact is based on possibility.", and $s_{cm}$ = "Yeh fact possibility par based hai.". It should be noted that in this example, $s_{cm}$ is a human-generated gold-standard code-mixed sentence.

As shown in Figure 3, the sentences are first preprocessed, i.e. converted to lowercase and punctuation is removed. All the English words in $s_{cm}$ are identified.

### 5.1.1 Word-Replacement

The next step is to translate $s_{cm}$ into X by individually translating all the English words detected in the sentence. However, the presence of homonyms (words which are spelled alike, but have different meanings) in $s_{cm}$ makes accurate translation difficult. For instance, the word 'rose' may refer to a type of flower, or 'to have got up', depending on the context. We handle this issue by translating the English words according to their PoS tags using an LLM. In order to get the PoS tags, we transliterate $s_{cm}$ in X[3], and translate this transliteration back to English to get a sentence $s_{temp}$. We then PoS tag $s_{temp}$, which is often not an accurate English translation of $s_{cm}$, but does allow for accurate PoS tagging. If $E_{cm}$ and $E_{temp}$ be the set of all English words in $s_{cm}$ and $s_{temp}$ respectively, then, ideally, since $E_{cm} = E_{cm} \cap E_{temp}$, we can translate all the English words in $s_{cm}$ into X according to their PoS tags. It should be noted that if any English words in $s_{cm}$ are not tagged due to discrepancies in $s_{temp}$, they are translated normally.

A sentence may also contain cross-language homonyms, i.e. English and X words which are spelled alike. In the example shown in Figure 3, the word 'par' is a cross-language homonym, as it means 'on' in Hindi and 'equal' in English. Naturally, it is a challenge to automatically determine whether the word is an English word, and thereby, whether it should be translated. To handle this issue, we consider words which occur more frequently in X as X words. We create for each language pair a dictionary $W_X$ of such words, such that $W_X = \{(w_1, t_1), (w_2, t_2), \ldots, (w_n, t_n)\}$ where $w_i$ is an English-X cross-language homonym, but is used more frequently in X, and $t_i$ is its transliteration in X. If X uses the Roman script, $t_i$ is the English translation of $w_i$. For example, if X is Spanish ('$es$'), the word 'soy' is a cross-language homonym, as it means 'am' in Spanish and refers to soya in English. As such, one of the $(w_i, t_i)$ pairs that $W_{es}$ contains is ('soy', 'am').

If a word $w_i$ in $s_{cm}$ is present in $W_l$, we replace it with $t_i$. In the example being discussed, since 'par' is used more frequently in Hindi than in English, it is replaced with its Hindi transliteration 'पर'.

### 5.1.2 Sentence Reconstruction

For the example shown in Figure 3, the sentence returned after word-replacement is "*yeh* तथ्य संभावना पर आधारित *hai*". This sentence is transliterated in X: "यह तथ्य संभावना पर आधारित है". Finally, we translate this X sentence into English to get an English sentence $s_{en}$: "this fact is based on possibility", which is a reconstruction of $s_r$ from $s_{cm}$.

The score $q$ returned by GAME is the semantic similarity between $s_r$ and $s_{en}$. In the example shown in Figure 3, $s_{cm}$ is correctly assigned the maximum possible score of 1.

---
[3]If X uses the Latin script, transliteration steps are skipped.

| CMD | Sentence | BLEU | GAME |
|---|---|---|---|
| 0.00 | Jaise hi bhor ki pehli kiranen aasman ko gulabi aur sone ke rang mein rangti, nidralu nagar jeevan mein aa gaya, ek naya din swagat karta hua, sambhavnaon aur khushiyon se bhara. | 47.85 | 97.39 |
| < 0.35[4] | Jaise hi **dawn** ki pehli **rays** aasman ko gulabi aur sone ke rang mein rangti, **sleepy** nagar jeevan mein aa gaya, ek naya din swagat karta hua, **possibilities** aur khushiyon se bhara. | 74.27 | 91.25 |
| 0.4 | Jaise hi **dawn** ki pehli **rays** aasman ko gulabi aur **gold** ke rang mein rangti, **sleepy** nagar jeevan mein aa gaya, ek naya din swagat karta hua, **possibilities** aur khushiyon se bhara. | 82.08 | 91.25 |
| 0.7 | Jaise hi **dawn** ki **first rays** aasman ko **pink** aur **gold** ke rang mein rangti, **sleepy** nagar **life** mein aa gaya, ek **new** din swagat karta hua, **possibilities** aur khushiyon se bhara. | 63.68 | 97.39 |
| 1.00 | Jaise hi **dawn** ki **first rays sky** ko **pink and gold** ke rang mein rangti, **sleepy town life** mein aa gaya, ek **new day** swagat karta hua, **possibilities and joys** se bhara. | 54.56 | 97.39 |

Table 3: GAME vs BLEU

**English sentence**: "As the first rays of dawn painted the sky in hues of pink and gold, the sleepy town came to life, welcoming a new day filled with possibilities and joys."

**Reference sentence**: "*Jaise hi dawn ki pehli rays aasman ko gulabi aur gold ke rang mein rangti, sleepy town life mein aa gaya, ek naya din welcome karta hua, possibilities aur khushiyon se bhara.*"

Table 6 shows examples of evaluation of gold-standard code-mixed sentences across all four language pairs using GAME, along with Reconstructed Sentence.

## 5.2 Assessment of GAME's Robustness

### 5.2.1 Evaluating Semantically Equivalent Code-Mixed Sentences

As discussed previously (1, 5), a given English sentence often has multiple unique but semantically equivalent code-mixed translations. A robust code-mixed evaluation metric must, therefore, assign the same score to all these sentences.

Table 3 shows a subset of twelve semantically equivalent code-mixed translations of an English sentence, generated through CG (4), which vary in their CMDs. Ideally, all the twelve code-mixed sentences should get equal or similar BLEU scores. However, we find that BLEU scores vary significantly (**47.85-82.08**), while GAME scores are contained in a smaller, more accurate range (**91.25-97.39**). The BLEU scores have a standard deviation of **11.49**, while only **2.64** for GAME.

We perform this test on a larger dataset, for which we choose 100 English sentences from the HinGE dataset (Srivastava and Singh, 2021), where each English sentence has 2-8 gold-standard Hinglish sentences corresponding to it. Similarly, we generate 57 English-Bengali and 62 English-Spanish generations for 15 and 21 English sentences respectively using CG. We evaluate the code-mixed sentences using both GAME and BLEU, and compute the average standard deviation of code-mixed sentences for each English sentence for both metrics. Average GAME score is 75.35 for CG English-Bengali sentences, and 92.63 for English-Spanish. Average standard deviations of GAME for Hinglish, English-Bengali and English-Spanish test data are 7.94, 6.50, and 1.71 respectively while for BLEU, these numbers are 43.53, 25.41 and 24.84. The lower standard deviation of GAME indicates that there is less variation in the scores it assigns to the equally correct (semantically equivalent) code-mixed translations of an English sentence, thus confirming that it is more robust than BLEU in this regard.

### 5.2.2 Extreme Case Evaluation

An effective code-mixed evaluation metric must assign high scores to quality translations and low scores to poor translations. In order to gauge GAME's accuracy in this area, we perform two tests on our own dataset:

(i) **Extreme Case 1**: We use GAME to evaluate our gold-standard code-mixed sentences. As the sentences are gold-standard and human-generated, the human score should be maximum.

(ii) **Extreme Case 2**: We use GAME to evaluate our gold-standard code-mixed sentences against an unrelated English reference sentence from the same dataset. Since the reference sentence and the candidate sentence are unrelated, the human score should be 0.

The results of this test are presented in Table 4.

| Language Pair | en-hi | en-fr | en-es | en-bn |
|---|---|---|---|---|
| Test Dataset Size | 188 | 221 | 320 | 521 |
| **Extreme Case 1** | 76.30 | 88.35 | 84.39 | 76.08 |
| **Extreme Case 2** | 7.69 | 6.88 | 7.92 | 7.06 |

Table 4: GAME scores (out of 100) for the test dataset.

## 6 Conclusion

We release a dataset containing gold-standard code-mixed sentences spanning 4 language pairs: English-{Hindi, Bengali, French, Spanish}. We propose Controlled Generation: a novel method for code-mixed text generation that allows for control over the Code-Mixing Degree of the genera-tion. We then show the inefficacy of the BLEU score in evaluating code-mixed text, and propose GAME as a solution. Our results show that GAME is more accurate and consistent than the BLEU score.

---
[4]We get the same sentence for $CMD \in \{0.1, 0.2, 0.3\}$. This is because for these values, $f(hi) = 0$. See 4.2.2

## Limitations

GAME can only handle homonyms that are different parts of speech. For instance, the word 'bat' may be translated to mean the flying mammal or a cricket bat. This could lead to inaccuracies in translation. Secondly, in some cases, there are slight inaccuracies in LID, transliteration, and translation. Using better-performing tools for these operations may help mitigate such issues. Due to budget constraints, we do not try using other LLMs such as GPT-4 (Brown et al., 2020) instead of Gemini Pro in GAME. Using better-performing LLMs or tools for the translation tasks may significantly boost GAME's performance.

Hindi, as discussed in section exhibit inflection in verbs. Bengali is also a highly inflectional language and thus, presents comparable challenges for the generation of English-Bengali as well. Although a similar approach can be employed for English-Bengali, we plan to explore it further in the future.

## Ethics Statement

The human annotators are volunteers paid commensurate to their efforts.

## References

Gustavo Aguilar, Sudipta Kar, and Thamar Solorio. 2020. LinCE: A centralized benchmark for linguistic code-switching evaluation. In *Proceedings of the Twelfth Language Resources and Evaluation Conference*, pages 1803–1813, Marseille, France. European Language Resources Association.

Emily Ahn, Cecilia Jimenez, Yulia Tsvetkov, and Alan W Black. 2020. What code-switching strategies are effective in dialog systems? In *Proceedings of the Society for Computation in Linguistics 2020*, pages 254–264, New York, New York. Association for Computational Linguistics.

Tom Brown, Benjamin Mann, Nick Ryder, Melanie Subbiah, Jared D Kaplan, Prafulla Dhariwal, Arvind Neelakantan, Pranav Shyam, Girish Sastry, Amanda Askell, et al. 2020. Language models are few-shot learners. *Advances in neural information processing systems*, 33:1877–1901.

Marine Carpuat. 2014. Mixed language and code-switching in the Canadian Hansard. In *Proceedings of the First Workshop on Computational Approaches to Code Switching*, pages 107–115, Doha, Qatar. Association for Computational Linguistics.

Hyung Won Chung, Le Hou, Shayne Longpre, Barret Zoph, Yi Tay, William Fedus, Eric Li, Xuezhi Wang, Mostafa Dehghani, Siddhartha Brahma, Albert Webson, Shixiang Shane Gu, Zhuyun Dai, Mirac Suzgun, Xinyun Chen, Aakanksha Chowdhery, Sharan Narang, Gaurav Mishra, Adams Yu, Vincent Zhao, Yanping Huang, Andrew Dai, Hongkun Yu, Slav Petrov, Ed H. Chi, Jeff Dean, Jacob Devlin, Adam Roberts, Denny Zhou, Quoc V. Le, and Jason Wei. 2022. Scaling instruction-finetuned language models. *arXiv preprint*.

Mrinal Dhar, Vaibhav Kumar, and Manish Shrivastava. 2018. Enabling code-mixed translation: Parallel corpus creation and MT augmentation approach. In *Proceedings of the First Workshop on Linguistic Resources for Natural Language Processing*, pages 131–140, Santa Fe, New Mexico, USA. Association for Computational Linguistics.

Ayush Garg, Sammed Kagi, Vivek Srivastava, and Mayank Singh. 2021. MIPE: A metric independent pipeline for effective code-mixed NLG evaluation. In *Proceedings of the 2nd Workshop on Evaluation and Comparison of NLP Systems*, pages 123–132, Punta Cana, Dominican Republic. Association for Computational Linguistics.

Devansh Gautam, Prashant Kodali, Kshitij Gupta, Anmol Goel, Manish Shrivastava, and Ponnurangam Kumaraguru. 2021. CoMeT: Towards code-mixed translation using parallel monolingual sentences. In *Proceedings of the Fifth Workshop on Computational Approaches to Linguistic Code-Switching*, pages 47–55, Online. Association for Computational Linguistics.

Deepak Gupta, Asif Ekbal, and Pushpak Bhattacharyya. 2020. A semi-supervised approach to generate the code-mixed text using pre-trained encoder and transfer learning. In *Findings of the Association for Computational Linguistics: EMNLP 2020*, pages 2267–2280, Online. Association for Computational Linguistics.

I-Hung Hsu, Avik Ray, Shubham Garg, Nanyun Peng, and Jing Huang. 2023. Code-switched text synthesis in unseen language pairs. *Preprint*, arXiv:2305.16724.

George A Miller. 1995. Wordnet: a lexical database for english. *Communications of the ACM*, 38(11):39–41.

Sneha Mondal, Ritika, Shreya Pathak, Preethi Jyothi, and Aravindan Raghuveer. 2022. CoCoa: An encoder-decoder model for controllable code-switched generation. In *Proceedings of the 2022 Conference on Empirical Methods in Natural Language Processing*, pages 2466–2479, Abu Dhabi,

United Arab Emirates. Association for Computational Linguistics.

Ravindra Nayak and Raviraj Joshi. 2022. L3Cube-HingCorpus and HingBERT: A code mixed Hindi-English dataset and BERT language models. In *Proceedings of the WILDRE-6 Workshop within the 13th Language Resources and Evaluation Conference*, pages 7–12, Marseille, France. European Language Resources Association.

Kishore Papineni, Salim Roukos, Todd Ward, and Weijing Zhu. 2002a. Bleu: a method for automatic evaluation of machine translation. pages 311–318.

Kishore Papineni, Salim Roukos, Todd Ward, and Wei-Jing Zhu. 2002b. Bleu: a method for automatic evaluation of machine translation. In *Proceedings of the 40th Annual Meeting of the Association for Computational Linguistics*, pages 311–318, Philadelphia, Pennsylvania, USA. Association for Computational Linguistics.

Tanmay Parekh, Emily Ahn, Yulia Tsvetkov, and Alan W Black. 2020. Understanding linguistic accommodation in code-switched human-machine dialogues. In *Proceedings of the 24th Conference on Computational Natural Language Learning*, pages 565–577, Online. Association for Computational Linguistics.

Braja Gopal Patra, Dipankar Das, and Amitava Das. 2018. Sentiment analysis of code-mixed indian languages: An overview of sail_code-mixed shared task icon-2017. *Preprint*, arXiv:1803.06745.

Shana Poplack. 2001. Code-switching (linguistic).

Thamar Solorio and Yang Liu. 2008. Learning to predict code-switching points. In *Proceedings of the 2008 Conference on Empirical Methods in Natural Language Processing*, pages 973–981, Honolulu, Hawaii. Association for Computational Linguistics.

Vivek Srivastava and Mayank Singh. 2021. HinGE: A dataset for generation and evaluation of code-mixed Hinglish text. In *Proceedings of the 2nd Workshop on Evaluation and Comparison of NLP Systems*, pages 200–208, Punta Cana, Dominican Republic. Association for Computational Linguistics.

Ruochen Zhang, Samuel Cahyawijaya, Jan Christian Blaise Cruz, Genta Winata, and Alham Aji. 2023. Multilingual large language models are not (yet) code-switchers. In *Proceedings of the 2023 Conference on Empirical Methods in Natural Language Processing*, pages 12567–12582, Singapore. Association for Computational Linguistics.

# A  Other Essential Supplementary for Controlled Generation

## A.1  Prompts for Base Creation (Language-Agnostic)

### A.1.1  Prompt A

> **Prompt A**
>
> For the given English sentence, do the following:
> 1. POS Tagging of the sentence
> 2. For the words which are either Noun (NN), Adjective (JJ), Adverb (RB), CC, or Interjection (UH), create a dictionary Imp_Eng
> 3. Translate the original English sentence into **MATRIX**
> 4. From Imp_Eng, look for the corresponding meaning in **MATRIX** and look them up in the **MATRIX** sentence. Create a dictionary **MATRIX**_eng_dict
> 5. Transliterate each **MATRIX** word in **MATRIX**_eng_dict in Roman in three ways or spellings and add that in the dictionary.
> 6. Format above as RFC8259 compliant json dictionary, in the format ["eng": <eng_word>, "pos_tag": <PoS Tag>, "**MATRIX**": <**MATRIX**_word>, "roman_**MATRIX**": <transliterations>]
>
> English sentence : text

- Here, **MATRIX** is the matrix language.

- Step 5 in above prompt can be modified for Spanish and French, as: "for each **MATRIX** word in **MATRIX**_eng_dict give three spellings that can be found in social media or twitter and add that in the dictionary."

- This works only for GPT-4 model and doesn't give the appropriate output in GPT-3.5-Turbo and Gemini-Pro

- The output is not just final dictionary and thus needs additional code or prompts to extract the sentence as well as dictionary (the base).

- Since, the model outputs all the steps and the steps have been mentioned in this manner, this gives the best results.

### A.1.2 Prompt B

> **Prompt B**
>
> For the given English sentence, do the following: create this RFC8259 compliant json dictionary in the format *{"hindi_trans": <hindi translation>, "Word_Dict":[{"eng":<eng word>, "base_eng":<base form of the english word>, "eng_pos_tag":<English PoS Tag>, "hindi":<hindi word>, "roman_hindi": <three different spellings of roman transliteration for hindi word>}]}*
>
> by doing PoS tagging of english sentence and then only choosing the words which are either Noun (NN), Adjective (JJ), Adverb (RB), CC, or Interjection (UH).
> And then translating the english sentence into Hindi and then looking for the corresponding meaning of these english words in that. Also, for each hindi word, transliterate it into three different spellings that can be seen in twitter.
> The output should be RFC8259 compliant json dictionary without any additional words or description
>
> english sentence : {Eng_sent}

- Here, Hindi is the matrix language. And thus **hindi** can be replaced with the other language for a different language pair.

- This works for GPT-4, GPT-3.5-Turbo as well as Gemini-Pro

- The output is the final dictionary which includes matrix language translated sentence as well and thus needs no prompts for further extraction.

- This works best for GPT-4. For Gemini-Pro, we have found some problems such as it does not give the asked spelling variations, and incorrect corresponding meaning of an English word in matrix language sentence, and occasional inclusion of the PoS tags that were not asked in the prompt. As for these PoS tagged words, they can be removed easily in the later part.

### A.2 English-Hindi specific Controlled Generation

#### A.2.1 Base Prompt for English-Hindi Specific CG

For this, GPT-4 has been used. This prompt is a modification of Prompt B.

> **Prompt for English-Hindi specific CG**
>
> For the given English sentence, do the following: create this RFC8259 compliant json dictionary in the format *{"hindi_trans": <hindi translation>,"Word_Dict":[{"eng":<eng word>, "base_eng":<base form of the english word>, "eng_pos_tag":<English PoS Tag>, "hindi":<hindi word>, "base_hin":<base form of the hindi word>,"hin_verb_type":<ACTIVE or PASSIVE or NA>, "roman_hindi": <three different spellings of roman transliteration for hindi word>}]}*
>
> by doing PoS tagging of english sentence and then only choosing the words which are either Verb, Noun (NN), Adjective (JJ), Adverb (RB), CC, or Interjection (UH).
> And then translating the english sentence into Hindi and then looking ofr the corresponding meaning of these english words in that.
> Also, for the english words that are verbs, check in the hindi sentence, if the respective hindi verb is active or passive, or if it isn't verb then 'NA'
>
> The output should be RFC8259 compliant json dictionary without any additional words or description
> english sentence : {Eng_sent}

The modification from Prompt-B gives the following new information:

- Base form of Hindi word

- Whether Verb is Active or Passive.

#### A.2.2 Rules for handling inflections in English-Hindi

In Hindi, only verbs need to be handled as such. The following describe the working for this inflection part:

**Cleaning:**
In this type of prompt, we get un-necessary words like 'was', as well as words with PoS tags that were not asked in the prompt. So, these are removed. Filtering PoS tags is simple, but as for the un-necessary words, the list is:

["This", "this", "is", "a", "am", "on", "In", "in", "are", "be", "the", "was", "were", "been", "have", "has", "had"]

Also, while not wrong as per rules, but there are some words that are kept in Hinglish as it is and not replaced with English counterparts. We remove them as well, though this is optional. The list for such words is:

['say', 'said', 'go', 'went', 'gone', 'come', 'came', 'tell', 'told']

**Identifying the suffix in Verb:**
The Base Prompt has been run using GPT-4. 'Base Hindi', i.e. base form of verbs results in words, majority of which have "ना" (na) at the end. For example, base form of 'khelta' (plays), 'khelti', etc. is 'khelna' If "ना" (na) is not the suffix in base form, the word can be replaced simply with English verb. Such Hindi words mostly act as an adjective. Using the base form of the verb, the suffix in the actual Hindi word can be identified. This can be done by removing "ना" from the base form, and then compare the rest with the actual Hindi word. For e.g. if the Hindi word is 'khelta' and as base form is 'khelna', by removing "ना" (na) from the base form, we have 'khel'. And if we remove 'khel' from the actual word, we get 'ta', which is the suffix that is necessary for the further steps.

**Word to be added after the English counterpart (Added_Word) :**

1. if suffix is '**ना**'

    - Such words will not be treated as a verb generally, and thus can be simply replaced with the English counterpart, except for the following subcase.

    - Subcase: If the sentence ends with this word, i.e. there is a connector 'और' (and) or '।' at the end, then **Added_Word** is **'करना'**. This is the word that will be added after the English counterpart.

2. if suffix is any of these - ["**ने**", "**नी**", "**ता**", "**ती**", "**ते**", "**ो**"]:

    - Subcase: if the verb is passive, then Added_Word='हो'+ suffix (Do not add suffix for the case "ो")
      E.g. if actual word is बदलने (badalne), English counterpart is 'transform', then the replacement will be 'transform होने' (transform hone), where 'होने' is the Added_Word.

    - Subcase: if the verb is active, then Added_Word='कर' + suffix

3. if suffix is '' i.e. there is **no** suffix

    - then Added_Word is 'कर', except for the following subcase.
    - Subcase: If there is already such added word in the Hindi sentence itself, i.e. if the next word is either from the list ['कर','करें', 'करता','करती','करते','करो', 'हुआ','हुए','हुई' ], then, Added_Word is kept blank ('').

4. if suffix is any of these - ['ा', 'े', 'या']

    - Subcase: if next word is either of ['हुआ','हुए']:, Added_Word is kept blank ('')
    - Subcase: if Hindi Verb is passive: Added_Word is 'हुआ' for suffix 'ा' and 'या', is 'हुए' for suffix 'े', with the following exception:
        - Sub-SubCase: if next word is either of ['गया', 'गई', 'गए'], then, Added_Word='किय'+suffix ('किया' if suffix is 'या')
    - Subcase: if none of above is the case, then Added_Word= 'किय'+suffix ('किया' if suffix is 'या')

5. if suffix is any of these - ['एं', 'े']:

    - Added_Word is 'करें'

6. if suffix is any of these - ['ी', 'ई'] :

    - Subcase: if next word is either of these ['हुई', 'की'], then, Added_Word is kept blank ('').
    - Subcase: if Hindi Verb is Passive, then Added_Word is 'हुई'
    - Subcase: If it is none of the above, then Added_Word is 'की'

The code has been included in the given link for the datasets, that implements all these steps.

### A.3 Quantitative Analysis of CG

A total of 8081 unique English-Hindi sentences were generated using Controlled Generation for 1840 English sentences randomly taken from HinGE dataset (Srivastava and Singh, 2021). Step 1 (4.2.1) was done using Prompt B (Section A.1). Despite not having the best performance, Gemini-Pro was used due to budget limitations. The generations as well as the reference sentences were in Roman script. While there were multiple code-mixed translations by Controlled Generation per English sentence, for calculating *Corpus BLEU*, one translation per English sentence was needed. So, for each set of multiple references and multiple generations, a pair was selected so as to get the maximum BLEU score.

Upon evaluation, the *Corpus BLEU* score achieved was 16.81

### A.4 Additional Information for Qualitative Analysis of CG

For a qualitative analysis of CG, we analyze the code-mixed sentences with varying CMD values for a set of 65 English sentences generated using Prompt A (Section A.1) with GPT-4 (Brown et al., 2020), the combination that gives the best outputs as per our observations.

### A.5 Controlled Generation with English as Matrix Language

| CMD | Sentence |
|-----|----------|
| 0.5 | The questions are of char types. |
| 0.7 | The questions are of char prakar. |
| 1.0 | The prashan are of char prakar. |

Table 5: Code-Mixed English-Hindi sentences generated using "Controlled Generation with English as Matrix Language".
English Sentence: "The questions are of four types."

For this, Step 1 remains the same. In Step 2, the algorithm remains same up to the step where the scores (ratio of the count of the English word to that of the Hindi word) for each replaceable word are calculated. Then, instead of replacing non-English (Hindi) chosen words with their English terms, English words in the English sentence are replaced with corresponding Hindi terms with the low scores 's' prioritized. This is because a low score implies low frequency of English word in the Real world dataset and high frequency for the non-English (Hindi) corresponding word and this means that the non-English (Hindi) term for the word is more commonly used than the English term.

## B Other Essential Supplementary for GAME

### B.1 Evaluation Using GAME, along with Reconstructed English Sentence

Table 6 gives examples of Evaluation of English-X gold-standard sentences using GAME.

### B.2 Test Dataset for Evaluation of GAME

For some words in certain sentences, Gemini Pro returns an empty response, due to which we omit these sentences in testing.

In second test of section 5.2,
For English-Bengali, 480 sentence pairs are the ones having Gupta et al. 2020 as English source and 41 are the ones that have been created by correcting and translating twitter dataset(Patra et al., 2018) to English.
For English-Hindi, 88 sentences are the ones having Gupta et al. 2020 as English source and 100 are the ones that have been created by correcting the twitter dataset(Dhar et al., 2018)
For English-Spanish, 269 sentences are the ones having Gupta et al. 2020 as English source and 51 are the ones that have been created by correcting and translating twitter dataset(Aguilar et al., 2020) to English.
for English-French, all 221 sentences are the ones having Gupta et al. 2020 as English source.

As for models, GPT-3.5-Turbo has been used only for English-Bengali and the twitter dataset part for English-Hindi. For others, Gemini-Pro has been used.

## C Additional Explorations

### C.1 Additional Explorations for Controlled Generation

#### C.1.1 Using Masked Language Modeling for calculating scores

We have used twitter code-mixed English-Hindi data for calculating counts of words and thus, the scores in controlled generation. According to us, we can improve the generations further if we consider context, and the position of word into account as well. According to us, it is very rare to find an uncommon (low count) Hindi word as well as an uncommon English word in same code-mixed sentence. Furthermore, for a sentence, in which there is a noun and its adjective, we think that if the adjective is in English, then the next priority should be given to the noun. For example, let's consider the English sentence "The questions are of four types" for which the Hindi translation is *"Prashn char prakaar ke hein"*. Here, *char* (four) is adjective for the noun *prakaar* (types). According to us, if four has been used in the code-mixed sentence, then, the next priority should be given to its noun, i.e. types should be used. We think that this makes the code-mixed sentences more natural. There can be exceptions though, but we think that presence of specific other words already present in the sentence may dictate the choices of replacement. We propose that this can be done if Masked Language Modelling (MLM) is done, where the model is trained on code-mixed dataset. For score of particular word, we mask the word in the sentence and then check the scores for the Hindi word and the corresponding English word in the possible choices for that masked position according to the model. For this, we experimented with HingMBERT (Nayak and Joshi, 2022). However, the outcomes were unsatisfactory. Frequently, the specific word for which we sought the MBERT score did not appear in the results for the masked word.

| | $s_r$ (English Sentence) | $s_{cm}$ (Code-Mixed Sentence) | $s_{en}$ (Reconstructed English Sentence) | $q$ |
|---|---|---|---|---|
| en-hi | Licensing and import policies were liberalized. | license tatha import ki policies ko udar banaya gaya | Licensing and import policies were liberalized | 100.0 |
| | The Congress should remove such differences, not create them. | congress ko inn differences ko remove karna chahiye inko create nahi | Congress should remove these differences, not create them | 92.05 |
| en-bn | But it sure makes you feel alive, doesn't it? | kintu eta surely tomake jibito feel korai ta ki na | But this must make you feel alive doesn't it | 88.25 |
| | Oh, I see you have already met my father. | oh ami dekhchi tumi amar babar sathe already meet korecho | Oh I see you have already met my father | 100.0 |
| en-fr | First it is certain that complete decoupling lowers production | tout dabord il est certain que le decoupling complet abaisse the production | first of all it is certain that complete decoupling lowers production | 97.97 |
| | The moment of truth is approaching | le moment de truth approche | the moment of truth is approaching | 100.0 |
| en-es | We voted against the report for the following reasons. | hemos votado en contra del report por los siguientes reasons | we voted against the report for the following reasons | 100.0 |
| | For example, we would have liked greater flexibility. | por example nos habra gustado greater flexibility | for example we would have liked more flexibility | 96.76 |

Table 6: Evaluation of English-{Hindi, Bengali, French, Spanish} gold-standard sentences using GAME

### C.1.2 Controlled Generation using 1-Shot Prompting

We also tried parameterized generation solely using prompting as well. We tried this for Hindi-English. The exact prompt has been shown here.

> **Prompt for Parameterized/Controlled Generation**
>
> Given a DCM value between 0 to 1, generate a code-mixed Hindi-English sentence in Roman for the given English sentence similar to the given example:
> **Example English sentence:** "This fact is based on possibility."
>
> DCM=0 (pure Hindi), then output: "Yah tathya sambhavna par adharit hai."
> DCM=0.25 (some English in Hindi), then output: "Yah fact sambhavna par adharit hai."
> DCM=0.5 (more English in Hindi), then output: "Yah fact possibility par adharit hai."
> DCM=0.75 (mostly English and some part Hindi), then output: "Yah fact possibility par based hai."
> DCM=1 (English only), then output: "This fact is based on possibility."
> Given DCM=0.25
> Given English sentence : "This trip is going to be difficult I guess"

As it can be seen in the prompt, we also gave description for the DCM values. For the DCM value equal to 0 or 1, the output is correct but for any other value, the output is the same code-mixed sentence, i.e. the model is not able to distinguish between different code-mixed sentences and thus we are not able to parameterize the degree of code-mixing. We tried this experiment on Gemini Pro, GPT-3.5-turbo as well as GPT-4, but this didn't work in any of the models.

## D  Additional Information

### D.1  Dataset Creation

#### D.1.1  Annotation

There were two annotators for the English-Hindi, English-Bengali, and English-French splits, and three for the English-Spanish split. We explained the phenomenon of code-mixing clearly to all the annotators. The human annotators, being bilingual and multilingual individuals who are fluent in English and the target language, had already engaged in code-mixing extensively. For each language pair, the annotators created gold-standard data independently.

Instructions to the annotators: "Please create a code-mixed sentence corresponding to the given English sentence"
The annotators were asked to ensure that the following criteria are met:
1. All code-Mixed sentences should be grammatically correct.
2. A code-mixed sentence should accurately convey the meaning of the corresponding English sentence, i.e. it should be a correct code-mixed translation of the English sentence.

The annotators participated voluntarily and they were paid commensurate to their efforts.

### D.1.2 Details of Datasets

For English-Bengali, 500 sentence pairs are the ones having Gupta et al. 2020 as English source and 41 are the ones that have been created by correcting and translating twitter dataset(Patra et al., 2018) to English.
For English-Hindi, 120 sentences are the ones having Gupta et al. 2020 as English source and 250 are the ones that have been created by correcting the twitter dataset(Dhar et al., 2018)
For English-Spanish, 294 sentences are the ones having Gupta et al. 2020 as English source and 53 are the ones that have been created by correcting and translating twitter dataset(Aguilar et al., 2020) to English.
for English-French, all 248 sentences are the ones having Gupta et al. 2020 as English source.

Also, Out of these, 145 English sentence pairs are common between English-Bengali and English-Spanish pair, making it parallel dataset across language pairs.

### D.1.3 Examples of Dataset

Table 7 shows the examples of English sentences and their corresponding gold standard code-mixed sentences.

### D.2 Controlled Generation
#### D.2.1 Models and Parameters
Gemini-Pro and GPT-4 have been used. The temperature was set to 0 for both cases with other parameters at default.

#### D.2.2 Computation
Not much computation is involved. Except for OpenAI API's use, it takes negligible time. Everything included, for processing and generating different Code-Mixed sentences for an English sentence, it may take upto 50 seconds depending on the API time.

### D.3 GAME
#### D.3.1 Models and Parameters
We use WordNet for identifying English words (Miller, 1995) in HinGE dataset evaluation. We use the Universal Sentence Encoder[4] to compute semantic similarity.

Spacy[5] ("en_core_web_sm") has been used to POS tag the sentence. Google Transliterate[6] has been used for transliteration task. Google Translate[7] (API) has been used for the first temporary translation which is used as an approximate way to identify the English words in the Code-Mixed sentence. We have used Gemini Pro, for tasks like word language identification, word translation and word translation with PoS tag given. Except for the case where the word translation is done along with providing the information of its PoS tag, the choice is subjective and doesn't require LLM. In English-Bengali, GPT-3.5-turbo has been used for the LID task. The temperature was set to 0. Also, the final English sentence which has been translated from the final transliterated (in matrix language) sentence is translated using Gemini Pro, but this is also a subjective choice and any other translation method can be used.

NLTK [8] library has been used for BLEU. For our purpose, Sentence BLEU has been used, with Smoothing Function[9]

#### D.3.2 Computation
The algorithm doesn't require large computation resources. Though, since some APIs have been used, this adds up to take a lot of time. On average, it takes 30 seconds to evaluate one sentence which we believe may be reduced further by 5-15 seconds with more optimized code. The CPU time is negligible and can take upto 4 seconds atmost.

---
[4]https://tfhub.dev/google/universal-sentence-encoder/4
[5]https://spacy.io/
[6]https://pypi.org/project/google-transliteration-api/
[7]https://pypi.org/project/googletrans/
[8]https://www.nltk.org/
[9]nltk.translate.bleu_score.SmoothingFunction().method4

| Language pair | English sentence | Gold-standard code-mixed sentence |
|---|---|---|
| English-Hindi | But the demands of the present are imperious | lekin present ki demands imperious hain |
| English-Bengali | I'm well aware of that, Raikes. | Ami oi bepare bhalo kore aware, Raikes. |
| English-French | We are not starting from scratch here | Nous ne partons pas de scratch ici |
| English-Spanish | This will become clear shortly in the IGC. | Esto will become claro shortly en el IGC |

Table 7: Examples of 4 English sentences and their corresponding gold standard code-mixed sentences for the English-{Hindi, Bengali, French, Spanish} language pairs.

### D.3.3 Other Information

For English-Bengali, we find that Gemini Pro makes frequent errors in LID, and considers most Bengali words to be English words. Therefore, for this language pair, we use GPT-3.5-turbo for the LID task, and omit no sentences.

## E  Algorithm for Controlled Generation

Algorithm 1 demonstrates the working of Controlled Generation algorithmically.

This pseudo-code also includes the algorithm for the steps describes inside the prompt which were previously described directly in the flowchart.

## F  Algorithm for GAME

Algorithm 2 shows the working of GAME algorithmically.

## Algorithm 1 Algorithm for Controlled Generation

1: **Input:** English Sentence, Code-Mixing Degree (CMD) Parameter, Hinglish Dictionary ('Word':counts)
2: **Output:** Hinglish Sentence
3: **Prompting with GPT-4:**
4: $translatedSentence \leftarrow translate(sentence, src = ``en'', dest = ``hi'')$
5: $impWords \leftarrow \{\}$
6: **for all** $word$ **in** $EngSentence$ **do**
7:     $pos \leftarrow getPartsofSpeechTag(word)$
8:     **if** word **is** "NN" **or** "JJ" **or** "RB" **or** "CC" **or** "UH" **then**
9:         $translatedWord \leftarrow GetCorrespondingWord(word, TranslatedSentence)$ ▷ Prompted to get the respective Hindi word from the translatedSentence
10:         $transliteration1 \leftarrow transliterate(translatedword, dest = ``roman'')$ ▷ Because of romanized words in Dictionary as dictionary is from Twitter data
11:         $transliteration2 \leftarrow transliterate(translatedword, dest = ``roman'')$
12:         $transliteration3 \leftarrow transliterate(translatedword, dest = ``roman'')$
13:         $impWords \leftarrow word, pos, translatedWord, transliteration1, transliteration2, transliteration3$
14:     **end if**
15: **end for** ▷ Further prompts were used to extract desired data from prompt output
16: **Word Scoring:**
17: $engCounts \leftarrow \{\}$
18: $hinCounts \leftarrow \{\}$
19: **for all** $item$ **in** $impWords$ **do**
20:     $eng[Count] \leftarrow getCounts(word, hinglishDictionary)$ ▷ Get the counts from Hinglish Dictionary for the english word
21:     **for all** $transliteration$ **in** $item$ **do**
22:         $hin_i[Count] \leftarrow getCounts(transliteration_i, hinglishDictionary)$
23:     **end for**
24:     $hin[totalCount] \leftarrow \sum_{i=1}^{3} hin_i[Count]$ ▷ We add the counts for the three possible romanized variations of the hindi word
25:     Calculate score for entry:
26:     **if** $hin[totalCount] == 0$ **then** $score = inf$
27:     **else**
28:         $score = \frac{eng\_counts[entry]}{\sum hin\_counts.values()}$
29:     **end if**
30: **end for**
    ▷ Sort entries in Imp_words by score in descending order:
31: sorted_words gets sorted(Imp_words.items(), key=lambda x: x[1], reverse=True) ▷ inf is considered the highest value and so, will come first
32: **Words Replacement:**
33: Replacement in Hindi Sentence (translatedSentence)
34: $words\_replaced \leftarrow 0$ ▷ Intitialized
35: Desired_words_replacement = int(CDM * len(Imp_words)) ▷ int gives integer
36: **function** REPLACEWORD($sentence, dest\_word, replacement\_word$)
37:     Replace $dest\_word$ in $sentence$ with $replacement\_word$
38: **end function**
39: **for** $i \leftarrow 0$ **to** $len(sorted\_words)$ **do**
40:     $remaining\_replacements \leftarrow \max(0, Desired\_words\_replacement - words\_replaced)$
41:     **if** $score_i = inf$ **then**
42:         $code\_mixed \leftarrow REPLACEWORD(code\_mixed, translatedword, word)$
43:         $words\_replaced \leftarrow words\_replaced + 1$
44:     **end if**
45:     **if** $remaining\_replacements = 0$ **then**
46:         **break**
47:     **else**
48:         **while** $remaining\_replacements > 0$ **do**
49:             $code\_mixed \leftarrow REPLACEWORD(code\_mixed, translatedword, word)$
50:             $words\_replaced \leftarrow words\_replaced + 1$
51:         **end while**
52:     **end if**
53: **end for**

## Algorithm 2 GAME

1: **Input:** Reference English sentence, Candidate sentence, X language code
2: **Output:** Quality score
3: **procedure** EVALUATE(reference, candidate, 'en', 'xy')
4:     reference, candidate $\leftarrow$ preprocess(reference, candidate)
5:     $transliteration_{temp} \leftarrow transliterate(candidate, dest = `xy')$
6:     $translation_{temp} \leftarrow translate(candidate, src = `en', dest = `xy')$
7:     $pos \leftarrow getPartsOfSpeechTags(candidate)$
8:     $words \leftarrow tokenize(candidate)$
9:     $ctr \leftarrow 0$
10:     **procedure** TRANSLATEENGLISHWORDS(candidate, words, ctr)
11:         **for** $word$ **in** $tokens$ **do**
12:             **if** $word$ **not in** common **then**
13:                 **if** lid(word) **then**     ▷ If the word is English
14:                     $ctr \leftarrow ctr + 1$
15:                     **if** $word$ **in** pos **then**
16:                         $translatedWord \leftarrow translate_{POS}(word, ``English'', ``X'', pos)$
17:                     **else**
18:                         $translatedWord \leftarrow translate(word, ``English'', ``X'')$
19:                     **end if**
20:                 **else**
21:                     $translatedWord \leftarrow word$
22:                 **end if**
23:             **else**
24:                 $translatedWord \leftarrow common.get(word)$     ▷ Fetch transliteration from dictionary
25:             **end if**
26:             $i \leftarrow words.index(word)$
27:             $words[i] \leftarrow translatedWord$
28:         **end for**
29:         $translated \leftarrow preprocess(words)$
30:         **return** $translated, ctr$
31:     **end procedure**
32:     $firstTranslation, ctr \leftarrow translateEnglishWords(candidate, words, ctr)$
33:     $firstTranslation \leftarrow preprocess(firstTranslation)$
34:     **if** $ctr > 0$ **and** $ctr < len(words)$ **then**     ▷ If the sentence is code-mixed
35:         $transliteration \leftarrow transliterate(firstTranslation, dest = ``X'')$
36:         $translatedSentence \leftarrow translate(transliteration, src = ``X'', dest = ``en'')$
37:         **if** $hasNonAlphanumeric(candidate)$ **then**
38:             $score \leftarrow 0$
39:         **end if**
40:         $score \leftarrow semanticSimilarity(reference, firstTranslation)$
41:         **return** $score$
42:     **else**
43:         $score \leftarrow 0$
44:         **return** $score$
45:     **end if**
46: **end procedure**